# ONTOLOGY BASED INFORMATION EXTRACTION FOR DISEASE INTELLIGENCE

Prabath Chaminda Abeysiriwardana[1], Saluka R. Kodituwakku[2]

[1]*Postgraduate Institute of Science, University of Peradeniya, SRI LANKA*
*Email: abeysiriwardana@yahoo.com*

[2]*Department of Statistics and Computer Science, Faculty of Science, University of Peradeniya, SRI LANKA*
*Email: saluka.k@gmail.com*

***Abstract:*** *Disease Intelligence (DI) is based on the acquisition and aggregation of fragmented knowledge of diseases at multiple sources all over the world to provide valuable information to doctors, researchers and information seeking community. Some diseases have their own characteristics changed rapidly at different places of the world and are reported on documents as unrelated and heterogeneous information which may be going unnoticed and may not be quickly available. This research presents an Ontology based theoretical framework in the context of medical intelligence and country/region. Ontology is designed for storing information about rapidly spreading and changing diseases with incorporating existing disease taxonomies to genetic information of both humans and infectious organisms. It further maps disease symptoms to diseases and drug effects to disease symptoms. The machine understandable disease ontology represented as a website thus allows the drug effects to be evaluated on disease symptoms and exposes genetic involvements in the human diseases. Infectious agents which have no known place in an existing classification but have data on genetics would still be identified as organisms through the intelligence of this system. It will further facilitate researchers on the subject to try out different solutions for curing diseases.*

## I. INTRODUCTION

Today there are many diseases which cause many fold harms to humans. Data based on them are published in web in different formats in different places of the web. This makes those data unusable in most of the time as well as in the most of the contexts.

In medical field, data relevant to diseases is huge. If these data can be extracted from different places and from different formats to a one place with same format and with particular subject focused, the data itself will become an easy content of information to refer. More the information about diseases that exists in digital form, better the understanding about disease, disease environment, and its cause and so forth. Scientists, researchers and inventors add content pertaining to diseases to the web that is of an immensely diverse nature. This disease information on the web is growing closer to a real universal knowledge base, with the problem of the interpretation of its true context. So there is a clear need for the disease information to become more logically assembled thus ensuring a semantic web for disease intelligence. The aim of introducing semantics into the disease information is to enhance the precision of search, but also enable the use of logical reasoning on the disease information in order to answer queries. Also when a logical structure is incorporated to this information it will become machine/computer readable as well as machine/computer processable, ensuring some kind of intelligence associated with this information.

Why this disease intelligence information is important to researchers, medical practitioners as well as to general public? Disease like AIDS, Dengue and H1N1 fever have their own characteristics changed rapidly at different places of the world and those characteristics (Ex: DNA patterns, symptoms of the disease etc.) reported by doctors at those places are not quickly available to other researchers / doctors in the other side of the world for reference. For example, if a researcher wants to analyze large number of sets of DNA patterns he may want to use his own set of data as well as other set of data given through other sources. If he manually searches the relevance and freshness of other sets of data, it will be a tedious and error prone task. Although he uses orthodox search engines they will only provide much larger set of information which is still hard to refer due to its largeness and unsatisfactory order. If machine can filter in relevance / meaningful, fresh, coherent and consistent data then his task of research will become much easier.

While introducing the concept of disease intelligence and showing the potential of its viability in

today`s medical field, the other focus of this research project is to show how this disease intelligence can be achieved. The methodology used to achieve the disease intelligence through web is based on ontologies created using OWL (Web Ontology Language) as well as evaluated by the reasoners available today. The ontology created here is named as disease ontology and it serves as a means to structure the disease domain.

So following are the main objectives of the study:

1. Find out a proper way to extract the information about rapidly spreading and changing diseases.

2. Make ontology to extract the information about those rapidly spreading and changing diseases using a proper web semantic [1] language.

3. Make information extraction and other natural language processing tools, key enablers for the acquisition and use of that semantic information.

4. Propose / lay a foundation for the Disease Intelligence System (DIS).

## II. SURVEY OF PRIOR RESEARCH

Disease intelligence is a new term introduced with this research and it is not yet discussed among other researchers in the world. But the subject discussed using this new term is widely supported by many other research areas of interest. Some are medical science, health science, gene related sciences (proteins, amino acids, nucleotides etc.) and to some extent business intelligence. These entire subject areas are based on some kind of ontology and are implemented using many kinds of ontology languages / vocabularies. Three most recently discussed technologies are SKOS [2], OWL [2] and RIF [2].

When considering the large, more complex and more logic based disease ontology; SKOS cannot be used due to following reasons: 1) It is not a complete solution 2) It concentrates on the concepts only 3) There is no characterization of properties in general 4) It is simple from a logical perspective, i.e., only a few inferences are possible.

Complex applications based on disease intelligence need following characteristics:

1. Objects should be able to identify with different URIs

2. There should be disjointedness or equivalence of the classes

3. Construction of classes should be possible with more complex classification schemes in addition to naming the classes. This strengthens the ability of a program to reason about some terms. For example, if Disease has resources A and B with the same properties on the DNA sequence, then A and B are identical.

In OWL, the behavior of properties such as symmetric, transitive, functional, inverse functional, reflexive, irreflexive etc. can be characterized. It concentrates on "taxonomic reasoning". So OWL is considered as the best language among those three languages which covers most of above characteristics efficiently. Web content developed using OWL has greater machine interpretability than that developed by XML, RDF, and RDF Schema (RDF-S) [3]. It also provides additional expressive power along with a formal semantics. The OWL is a semantic web language designed by W3C Web Ontology Working Group [4] on World Wide Web consortium to represent rich and complex knowledge about things, groups of things, and relations between things. Other ontologies can refer these ontologies as well as these ontologies can import some other ontology to be fused with them. Also OWL is a part of the W3C's Semantic Web technology stack, which includes RDF [RDF Concepts] and SPARQL [5]. Ontologies developed with OWL contain objects. An object designated by a URI becomes information object "on the web". Objects destined to have URIs are also known as "First Class Objects" (FCOs) [6]. Tim Berners-Lee [7] has suggested that the Web works best when any information object of value and identity is a first class object. The most recent version of OWL is OWL 2 [8] and it has been used to form disease ontology.

There are some well-known medical vocabularies based on ontologies. They are complete to the extent that researchers, medical practitioners and general public can interact with them to extract information. They have been developed and continuously being developed for so many years by domain experts. Some of them are discussed here for the purpose of introducing the strong characteristics of them to the disease ontology while to eliminate weak characteristics being introduced into disease ontology.

SNOMED CT (Systematized Nomenclature of Medicine-Clinical Terms) [9] is considered to be the most comprehensive, multilingual clinical healthcare terminology in the world. It is a resource with comprehensive, scientifically-validated content. It contains electronic health records and a terminology that can cross-map to other international standards. It is already used in more than fifty countries. SNOMED CT has a hierarchy consists of more than 311,000 concepts pertaining to Electronic Health Records (EHR) and forms a general terminology for it. Several software applications are able to interact with it to extract the required information. This information is known to produce relevant information consistently, reliably and comprehensively as a way of producing electronic health records. The concepts are organized in hierarchies, from the general to the specific. This allows very detailed ("granular") clinical data to be recorded and later accessed or aggregated at a more general level. Concept descriptions [9] are the terms or names assigned to a SNOMED CT concept. There are almost 800,000 descriptions in SNOMED CT, including synonyms that can be used to refer to a concept.

The ontology used for SNOMED CT basically covers the clinical aspects of the disease domain. For example, SNOMED CT can be used to analyze how many cancer surgeries are performed and to consistently record outcome data to determine whether surgery has an impact on long-term survival and local recurrence in cancer treatments. But it does not give clues about some patients with special genetic sequence in their body to be able to quickly recover from cancer. This is because of the reason that genetic information is not considered in this ontology. Basically it uses the patients` clinical records and drugs used for those patients. The intelligence associated with this system is basically on how drugs affect the disease and how patients react to some drugs based on different conditions such as sex, age and may be genetics. Micro level analysis of gene in relation to patient and disease is not covered. So this system lacks the following details: the micro and macro level structure of the organisms (if it is an infection) which causes the disease, the DNA / RNA details of the patients etc. So the intelligence regarding to genetic side is not properly covered by this system. The proposed disease ontology is supposed to cover all these areas including clinical aspects and so the disease ontology is expected to integrate all these aspects. Further the disease ontology is expected to have the ability to import SNOMED CT ontology on to disease ontology to make the web of data for disease intelligence.

When micro level information of the disease ontology is considered, the prominent existing research work in relation to genes and genetic materials can be found at the GO (Gene Ontology) [10] Consortium. It can be considered as a virtual meeting place for biological research communities actively involved in the development and application of the Gene Ontology which consists with a set of model organism and protein databases. The Gene Ontology (GO) Consortium is established with the objective of providing controlled vocabularies to describe specific aspects of gene products. Collaborating databases annotate their gene products (or genes) with GO terms. These GO terms have references and indicate what kind of evidence is available to support the annotations. It is possible to make unique queries across databases as it uses of common GO terms. The GO ontologies have their concepts specialized to impart different level of granularity where attribution and querying can be performed according to necessities. The GO Vocabularies [10] are dynamic since knowledge relating to gene and protein roles in cells are continuously introduced and changed by the users.

There are three structured controlled ontologies in relation to gene products considering the biological processes, cellular components and molecular functions in a species-independent manner. So it is a kind of complete ontology in relation to gene products behave in a cellular context but not an ontology based on genetic aspect of organisms both humans and infectious agents. So it only covers the part of micro level profile of the disease ontology. Also it lacks the clinical aspect of the disease intelligence. So bridging information between diseased and infectious agent is not clearly covered by these ontologies to give clear cut evidence about disease intelligence.

Following GO, 150 Open Biomedical Ontologies (OBO) [11] are listed at the National Center for Bio-Ontology (NCBO) BioPortal. Those ontologies deal with molecular, anatomical, physiological, organismal, health, experimental information. But up to now with 20 different terms for "protein" associated with different ontologies it can be found that significant overlap exists with those ontologies. OBO Foundry promotes a set of orthogonal ontologies developed over basic categories drawn from the Basic Formal Ontology (BFO) [12] and encourages the reuse of basic, domain-independent relations from the Relational Ontology (RO) [12]. Here it is necessary to use well defined relations and make it clear when the relations are to be used, and what inferences, if any, may be drawn from them. So it is expected to remove such overlaps through disease ontology as it is built with broad spectrum of information in mind. Also categories drawn from the Basic Formal Ontology (BFO) and reuse basic, domain independent relations from the Relational Ontology (RO) will help the disease ontology more powerful in its context.

The HCLS (Health Care and the Life Sciences) [13] is a knowledge base where a collection of instantiated ontologies can be found. For example, interesting molecular agents can be found for the treatment of Alzheimer disease.

Relating to diseases there is one such important knowledge base implemented using ontologies called Pharmacogenomics Knowledge Base (PGKB) [14]. It contains logically arranged data to represent how genetics plays a role in effective drug treatment. It offers depression related pharmacogenomic information that facilitates additional knowledge curation beyond the PharmGKB database. Thus, ontologies like PharmGKB can play an important role in semantic data integration and guide curation activities with well-established use cases towards populating a specialized knowledge base. The disease intelligence covers this part as well and the disease ontology will be instrumental in achieving better treatments than expected from HCLS knowledge base.

One of the researches carried out and presented in the 2008 International Conference on Bio informatics & Computational Biology (BIOCOMP'08) [15] is annotating the human genome with Disease Ontology. In this research it says that the human genome has been extensively annotated with Gene Ontology for biological functions, but minimally computationally annotated for diseases. This research tries to evaluate the mapping of existing genome data with the existing disease ontologies. But such a mapping lacks the power of intelligence which may only be formed by considering genetics relevant to humans specially extracted through diseased humans through clinical records of diseased etc. The subtle differences recorded within patients' records may give vital clues regarding remedial or preventive measures for those diseases. The above annotations will only pave way to identify defected human genes or responsible genes associated with the diseases. But it does not discover the genes which may show resistant to some diseases as there is no mechanism to compare sufficient clinical records of patients in such a mapping. Also it is not going to consider and compare genetics associated with diseased and genetics associated with infectious agents. This cited research basically considers genetically based diseases (genetic disorders) but not the diseases caused by infectious agents. So it considers only a specific domain of disease ontology and human genome. The disease ontology discussed in this research covers more general and widely covered disease ontology which would be the minimum need for disease intelligence.

Another interesting ontology-based system is MeSH (Medical Subject Headings)[16] which has been listed as one of the prominent project under U.S. National Library of Medicine - National Institutes of Health and shows some relevance in regard to disease intelligence. It is a vocabulary thesaurus consists of sets of terms and naming descriptors in a hierarchical structure. This hierarchical structure permits searching of these terms at various levels of specificity. MeSH descriptors are arranged in both an alphabetic and a hierarchical structure. It has concepts called 'Headings'. At the most general levels of this hierarchical structure there are broad headings such as 'Anatomy' and 'Mental Disorders'. At the deep of the hierarchy more specific headings can be found. For example, at the twelve-level of the hierarchy, headings such as 'Ankle' and 'Conduct Disorder' can be found. There are 26,142 descriptors in 2011 MeSH. There are also over 177,000 entry terms that assist in finding the most appropriate MeSH Heading, for example, "Vitamin C" is an entry term to "Ascorbic Acid."

There are other kinds of thesaurus as well. One of the very well established thesaurus/ knowledge shelf with fascinating search capabilities is PubMed by U.S. National Library of Medicine, National Institutes of Health. PubMed [17] is a knowledge base comprising over 20 million citations. It refers to another knowledge base called MEDLINE [18] comprising life science journals, and online books. PubMed has citations and abstracts for the fields of medicine, nursing, dentistry, veterinary medicine, the health care system, and preclinical sciences. PubMed facilitates its users to access additional relevant Web sites and links to the other NCBI molecular biology resources.

The mechanism use to populate MEDLINE is associated with its forum where publishers of journals can submit their citations to NCBI and then they are allowed to access the full-text of articles at journal Web sites using LinkOut [19].

It is very important to have thesaurus in regard to disease intelligence as some information may come associated with some general terms as it is with Vitamin C which has a more scientific name 'Ascorbic Acid'. Disease information may come from patients themselves to these ontologies as some patients' record their experience associated with the disease they suffer and those record data may be incorporated to the disease intelligence under separate concept.

When drug details are incorporated into disease intelligence it can be expected to have information of drugs relating to drug usage and so drug business is coming under the purview of Disease Intelligence resulting in some kind of business intelligence revolve around it. But invoking business intelligence is not one of main concerns of making this disease ontology rather it would be allowed to automatically be sprung through the ontology with existing concepts.

So the disease ontology discussed in this research is a kind of universal ontology focused on disease intelligence.

### III. METHODOLOGY

When considering the disease intelligence, it is evident that the ontology based information extraction would be a promising niche of achieving disease intelligence.

Two well experimented approaches to make ontologies are the bottom up and top down approaches. Bottom up approach is not considered here to make disease ontology as this ontology is viewed in more general at the beginning and then more details are covered by concepts at the end. As multiple inheritance can be achieved and checked through reasoning techniques applied to the ontology (developed using OWL 2 language) while it is in the developing stages, top down approach has more advantages over bottom up approach.

The top-down approach is followed in modeling the domain of diseases. So the concepts developed at the beginning are very generic. Subsequently they are refined by introducing more specific concepts under those generic concepts. At some stages it seemed that a middle-out approach best suited for the purpose. At those stages much concern was focused to identify the most important concepts which would then be used to obtain the remainder of the hierarchy by generalization and specialization.

Several research groups have proposed some methodologies that can be applied in the development process of ontologies. Also there is no one 'correct' way or methodology for modeling the domain of interest using ontologies. Some of the methodologies used for ontology engineering are Skeletal Methodology [19], competency questions [20] (the questions will serve as the litmus test later), top-down or bottom-up or combination of both development processes [21], KACTUS [22], Methontology [23] and Formal Tools of Ontological Analysis [24]. But ontology engineering is still a relatively immature discipline so any development cycle is not hundred percent guaranteed for optimal results. Skeletal Methodology shows some success in building huge ontologies. Uschold et.al used this approach to create an Enterprise Ontology [25]. The TOVE [26] (TOronto Virtual Enterprise) project from University of Toronto`s Enterprise Integration Laboratory has developed several ontologies for modeling enterprises by using this approach. So this approach is used to build the proposed disease ontology.

1. Identify purpose – Clarify goal and intended usage of the ontology.

The disease ontology is to lay foundation for Disease Intelligence System by extracting information about rapidly spreading and changing diseases. Here every aspect related to human diseases is supposed to interconnect within one domain of interest which is the disease domain.

2. Building the ontology – This is broken down into three steps:

a. Ontology capture – Here key concepts and relationships are identified in the domain of interest. Precise unambiguous text definitions are created for such concepts and relationships and terms are identified to refer to them. A middle-out approach is used to perform this step, so identify the most important concepts which will then be used to obtain the remainder of the hierarchy by generalization and specialization. The most generic concepts considered here are Disease, DiseaseArea, DiseasePrevention, DiseaseStructure and DiseaseSymptoms. Huge amount of data related to these concepts already exists. But they lack following features to be considered as having capacity to generate disease intelligence.

1. Data related to different area of interest in disease domain is not interconnected.

2. Data is not logically arranged to be processed by machine.

It is necessary to interconnect key sub areas of disease domain and connect them and their data logically within the proposed disease ontology. A few key relationships used to interconnect those key areas are hasStructure and hasSymptoms.

b. Coding – Represent the knowledge acquired in 2.a. in a formal language - OWL2.

c. Integrate existing ontologies – Proper integration of other ontologies to this ontology is not implemented as such an activity needs the disease ontology to be further developed with more sub-concepts.

3. Evaluation – Make a judgment of the ontologies with respect to a frame of reference which may be the requirement specifications or competency questions.

The disease ontology is validated by testing it with the Protege 4.1 beta version (Open Source) developed by research team at The University of Manchester and Stanford. The fact plus plus (fact ++) plugging imported to Protege 4.1 software will act as a reasoner.

4. Documentation – Document ontologies according to the type and purpose. The documentation part of the disease ontology is not considered yet. But Entity annotations (human-readable comments made on the entity) are implemented to some extent.

Fact plus plus plugging which is used as the reasoner converts the asserted model for disease ontology into inferred model. Inferred model contains the disease information which are not explicitly stated in the disease ontology, but inferred from the definition of the disease ontology such as multiple inheritances.

At last a web site was developed from the ontology. The web site is a machine generated one with easy browsing capabilities. It represents the knowledge database of the proposed disease ontology and acts as a graphical user interface which facilitates the easy reference of disease intelligence information.

Following naming convention is used. Class names are capitalized and when there is more than one word, the words are run together and capitalize each new word. All class names are singular. Properties have prefix "has" or "is" before property name or "of" after property name when a verb is not used for a property. All properties begin with a simple letter and when there are more than one words, words are run together with capitalized first letter from second word. This prefixes and suffices further enable the intent of the property clearer to humans, as well as make its way into the "English Prose Tooltip Generator". It makes the tool acts as a natural language processing key enabler in this regard.

To determine the scope of the proposed disease ontology, a list of questions is sketched. This questionnaire should able to be answered by the knowledge base based on the proposed disease ontology. Following competency questions were initially put into be answered.

1. What cause a disease?
2. How can a disease be identified?
3. Is there any cure for a disease?
4. What is the relationship between cause of the disease and human body?
5. Does the organism have a particular attack site of the human body?
6. What kind of structure initiates a particular disease?
7. Whether some micro level structure of human body resists to some disease than some other structures?
8. Does genetics have high role in disease control?
9. How much does a drug affect the disease?
10. What structure or functionality of drugs is more effective on disease?
11. Is there any environmental impact on disease spreading?
12. Does a disease have a special affinity for particular human body part?

Based on the above questions initial class structure is built. While developing the class structure which has satisfactory answers to the above questions, reality of the disease ontology is considered as well. The proposed disease ontology is a model of reality of the existing diseases in its environment and the concepts in the ontology reflects this reality. Therefore, care is taken to build most generic six classes to reflect that reality.

To answer first competency question, a class namely Disease is built to store types of diseases. It has two categories of diseases represented by two sub-classes named as Autoimmune and Infectious. The answer to second competency question is generally by disease symptoms and specifically by diagnosis methods, so a class called DiseaseSymptoms is built to store data about disease symptoms and results obtained from disease diagnosis tests. Disease's symptoms are used for normally identify the disease and diagnosis results are used to confirm that the disease exists. These classes may be used by different types of users such as patients and doctors. The third question can be answered by building a class called DiseasePrevention which can be used to store data about the preventive methods and measures of diseases. Question five can be tackled with building a class called DiseaseArea where human body parts, vulnerable relating to particular disease, are described. DiseaseStructure class answers the question six as the class is supposed to store data about structure of disease area and structure of the infectious organisms. Question eight is answered by the class GeneticMaterial as it is supposed to provide place for storing genetic information about infectious organisms as well as human genetics.

The resulting base disease ontology has 6 most generic classes or concepts namely Disease, DiseaseArea, DiseaseSymptoms, DiseasePrevention, DiseaseStructure and GeneticMaterial. The root class of these six classes of the Disease ontology is the *Thing* class. OWL classes are interpreted as sets of individuals (or sets of objects). The class Thing is the class that represents the set containing all individuals. Because of this all classes are subclasses of Thing. The proposed Disease ontology has the following tree structure shown in Figure 1.

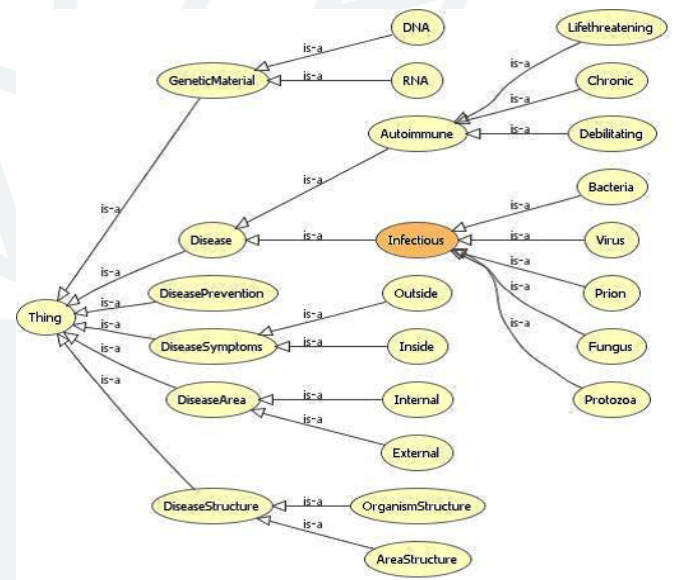

*Figure 1: Class Hierarchy of the Disease Ontology*

In modeling the interconnections between these classes, other questions play a vital role. Answer to the question four can be found by relating Disease class with DiseaseSymptoms class as cause of the disease can be found only by its symptoms on human body. Thus Disease class is interconnected with DiseaseSymptoms by hasSymptoms relationship. So this relationship makes it possible to map disease symptoms to disease. The question seven and eight can be answered by building the hasGenetics relationship between GeneticMaterial class and DiseaseStructure class.

The question nine and ten can be answered by building the hasPrevention relationship between Disease class and DiseasePrevention class. Because of transitive nature of the ontology the relationship between DiseaseSymptoms class and DiseasePrevention class can be built so drug effects on disease symptoms can be evaluated. The answer to question twelve forms the hasArea relationship between Disease class and DiseaseArea class. Human involvement with those diseases, other causes for human diseases and fine structure associated with those diseases both with regard to humans and other causative agents thus can be extracted through making relationships between these six classes.

## IV. RESULTS AND DISCUSSION

The following few paragraphs describe special characteristics which can also be noted in this proposed ontology.

The Disease class has the most important place in this ontology and it is named with the intention that other disease ontologies exist in the web may be imported to this ontology in future. It contains all the information regarding the disease with respect to origin of the disease and its data are logically arranged. It has two sub classes called infectious and autoimmune. Under infectious the diseases related to five most common infectious organisms/ agents namely Virus, Fungus, Prion, Bacteria and Protozoa are placed as sub classes. Micro level details of organisms related to diseases can be incorporated under OrganismStructure subclass by creating the relationship hasStructure between DiseaseStructure class and Disease class. It is a unique feature of this ontology because this disease ontology has molecular level details of both humans and most of other organisms. If someone wants to add another category of disease originated by an organism, it is not a difficult task to add it. The other sub class of the Disease class is the Autoimmune class and it has three sub-concepts/ classes called Debilitating, Chronic and Lifethreatening. Among these three classes, the most successful candidate for having other ontologies incorporated into it is Chronic class. The reason behind this is the mostly discussed topic among these three categories on the web is chronic disease. This is not validated by any research but just looking up through search engine may give hint about this end.

Then the class DiseaseStructure has two sub concepts/ classes called AreaStructure and OrganismStructure. AreaStructure class is for describing the affected area of the disease. Here only the details regarding structural changes at cellular level and below (molecular and sub molecular level) and functional changes are stored. So it shows clues about what kind of disease occur in that place. There may be a controversial issue regarding placing such a kind of class here and separate DiseaseArea class. But at this research it is thought that there exist some subtle and vital difference between those two classes and it is better to have them as separate classes rather than as a single class. Once ontology has been fully developed, the two classes can be merged, without difficulty. The reason behind the class to be allowing to be existed as a separate class is that it provides a unique way to represent micro level details of the humans in separate place. It is not necessary to identify the disease to place such details in this class as it is not directly derived from Disease class. The OrganismStructure contains details about the organism structure both in micro and macro level. Even the details available about organisms which are not yet associated with disease would also be placed with this class.

DiseaseArea has two sub classes called Internal and External. Internal has details about diseased internal parts of human body describing the internal parts both with respect to disease and not with respect to disease, if disease details are not available. This is basically about the human body parts and not the micro structure of the disease area. This identifies where the disease attacks and how sensitive the disease to that particular area of human body. Even statements given by patients about those areas can be stored here. So this class can be considered as some general class to store information about the disease. External class is same, except that it discusses external body parts of humans such as surface of skin, limbs, face, and hair and so on. Internal class and External classes are not disjoint as some parts may be discussed both in Internal and External classes.

DiseaseSymptoms class is responsible for explicitly storing whatever symptoms there regarding a disease. In real world, cause of the disease can be found only by its symptoms on human body. Disease class interconnected with DiseaseSymptoms class (by hasSymptoms relationship) makes it possible to map disease symptoms to disease. Sometimes disease may not be known but identify the abnormality in body as a kind of disease symptoms. So this class which is not under Disease class and act as separate class facilitates information regarding such kind of symptoms to make its way through to this class. DiseaseSymptoms has two classes called Inside and Outside. They are responsible for symptoms of inside and outside of the human body respectively.

Other class is DiseasePrevention and it contains information regarding disease prevention. It will have most results out of research work carried out by doctors, scientists, researchers, individuals etc. all over the world about disease prevention. Transitive nature of the ontology on the relationship between DiseaseSymptoms class and DiseasePrevention class allow the drug effects to be evaluated on disease symptoms. Also it will have highest portion of the information with the involvement of the reasoner where new relationships between drugs, patients` clinical records, trials on disease prevention etc. will be discovered to get new profile on disease intelligence. All other classes provide support to achieve this end of the disease intelligence.

The last of the most general classes is the GeneticMaterial class. It has details of DNA and RNA stored in DNA and RNA sub classes respectively. These classes are associated with Infectious and OrganismStructure classes through object properties. Because the GeneticMaterial class has its own separate class hierarchy; it can store more genetic information about organisms which have not yet reference to infectious disease.

Some classes are made explicitly disjoint here. In Infectious class, all subclasses are made disjoint to each other as no organism is fall into more than one class in this domain, i.e. the Infectious class cannot have any instances in common. The same is done for subclasses of Autoimmune, subclasses of DiseaseArea and subclasses of DiseaseStructure.

The proposed Disease ontology has some notable properties / slots / relations. Two of them are hasStructure and hasSymptoms with inverse properties isStructureOf and isSymptomsOf respectively. Although storing the information 'in both directions' or with inverse properties is redundant from the knowledge acquisition perspective, it is convenient to have both pieces of information explicitly available. This approach allows users to fill in the Disease in one case and the DiseaseStructure in another. When disease is not known disease structure can still be stored and described in relation to unknown disease. Also the knowledge-acquisition system could automatically fill in the value for the inverse relation ensuring consistency of the knowledge base, if the other value exists.

There are sub properties as well in the proposed Disease ontology. The hasOrganismStructure is a sub property of hasStructure. The hasAreaStructure is a sub property of hasStructure.

The proposed Disease ontology has defined domains and relevant ranges as well. For example, the domain and range for the hasSymptoms property are Disease and DiseaseSymptoms classes respectively. The domain and range for isSymptomsOf is the domain and range for hasSymptoms swapped over. Although the domains and ranges of hasSymptoms and isSymptomsOf properties are specified, it is not advisable to doi it over other properties of the Disease ontology without further studying those properties and classes covered by them. The reason behind this is that domain and range conditions do not behave as constraints. So they can cause 'unexpected' classification results which lead problems and unexpected side effects.

Also the proposed Disease ontology has restrictions. If a disease is there, at least a symptom should be there to indicate that the disease exists. Here an 'existential restrictions' is used to describe individuals in Disease class that participate in at least one relationship along a hasSymptoms (some) property with individuals that are members of the DiseaseSymptoms class. These restrictions are applied to the properties depicted by the dotted arrows in Figure 2.

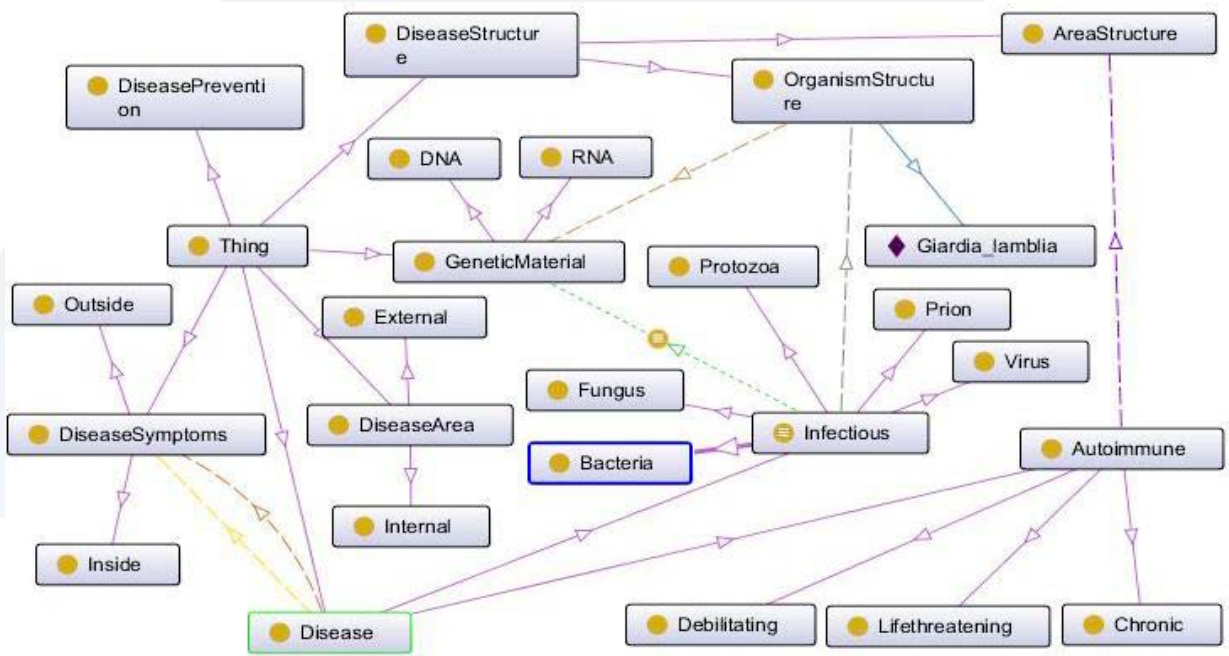

*Figure 2: Class Hierarchy with Properties of the Disease Ontology*

The proposed Disease ontology has primitive classes as well as defined classes to enable the reasoner to classify the ontology. One such defined class is Infectious and its icon has 3 horizontal lines on its orange sphere as depicted in Figure 2. This class enables necessary and sufficient condition for hasGenetics (hasGenetics some GeneticMaterial) object property and makes the class falls under equivalent classes. So when class is read with genetic material it will be classified under Infectious class. This has significant consequence in this disease ontology as some infectious agents can exist within known classification but still can be identified as organism because of available data on genetics.

The proposed Disease ontology has individuals in its classes. For example, OrganismStructure class under DiseaseStructure class has the individual *Giardia lamblia* with data property 'locomotion' with the value 'Flagellates'. So this individual has some relation to a disease which is an individual assigned to the Disease class and so acquired by the hasOrganismStructure sub-relationship between the disease and the organism. To make this individual uniquely identified, it is given a URI: 'http://www.disintel.lk/ontologies/disease.owl #Giardia_lamblia'. It should be noted that all the members of the OrganismStructure class are also the members of other super classes of it namely DiseaseStructure and Thing.

OrganismStructure class should be used to populate the proposed disease ontology with millions of organisms existing in the world either by importing ontologies which contain those individuals or adding those individuals by communities under the OrganismStructure class.

If the Disease ontology designed here is used to assist in natural language processing of articles in healthcare, health research and medical magazines / journals, it may be important to include synonyms and part-of-speech information [27] for concepts in the Disease ontology. This is little bit discussed when naming conventions are discussed. In addition to that, annotation which can be incorporated with the concepts will facilitate this.

Then this ontology should be made available through the web for ontology navigation. The rough interface generated for the ontology is shown in Figure 5. In other words, this shows the machine level understanding of the ontology so it is the intelligence that can be expected from the system.

The web site has components divided into logical groups and rendered in a linear fashion. So taking the Disease class for example, its enumerations if any, that is, intersectionOf (Closure axioms are used here for describing the genetics of the individuals of Infectious class), unionOf, and so on are listed in one group, while the properties related to it (through domain or range) are listed in another group. Standard description logic (DL) operators are used whenever they occur in class expressions to make the representation more clear and concise.

All entity references are represented by hyperlinks using unique URIs as the identifiers. Thus, clicking on an entity link in a particular document causes the view to shift directly to the linked entity's document. This is in keeping with the look and feel of traditional web-like viewing and navigation of documents. The evaluation of the disease ontology is done using the fact plus plus (fact ++) plugging imported to Protege 4.1 software. The fact plus plus (fact ++) plugging acts as a reasoner and validate the ontology against the logic base reasoning for discrepancy in multiple inheritance etc.

Inferred model shows that there are multiple inheritances associated with Infectious class for the disease ontology. This can be viewed by the Figure 3 where asserted and inferred class hierarchies are positioned side by side. The description of Infectious is converted into a definition and icon in front of Infectious class bears three horizontal white lines to indicate that it is a defined class. So if something is an Infectious then it is necessary that at least one genetic material (DNA or RNA) that is a member of the class GeneticMaterial is there. Moreover, if an individual is a member of the class OrganismStructure then it has at least one genetic material that is a member of the class GeneticMaterial. Then these conditions are sufficient to determine that the individual must be a kind of disease so it becomes a member of the class Infectious. This multiple inheritance has been automatically inferred by the reasoner as shown in Figure 3 and as a result, inferred model has OrganismStructure reclassified under Infectious class.

The reasoner also checks semantic consistency of the disease ontology such as satisfactory of the concepts or correctness of the concept hierarchy. The descriptions of the classes (conditions) are used to determine if super-class / subclass relationships exist between them. Here the reasoner tests whether a class is a subclass of another class or not (subsumption testing). In this testing, the proposed Disease ontology shows that the inferred class hierarchy has no consistency problem with respect to the asserted class hierarchy as reasoner doesn't show any warning sign or red colored class names in Figure 3.

So it indicates that the expected class hierarchy has no discrepancy in its design.

Other testing done using the reasoner is consistency checking of the disease ontology. Based on the conditions of classes in the disease ontology the reasoner checks whether or not it is possible for classes to have any instances.

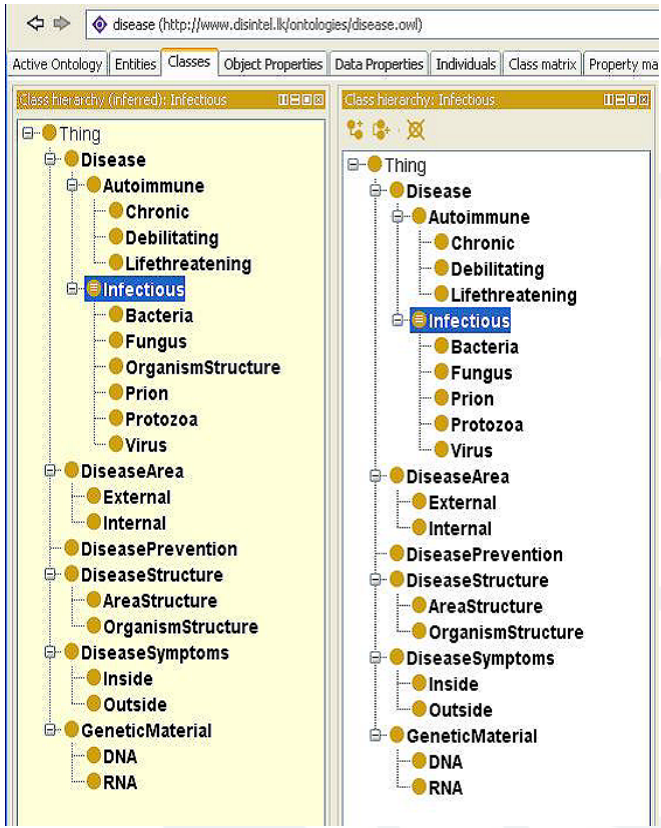

*Figure 3: Asserted and Inferred Class Hierarchies*

According to this testing on the disease ontology, each and every class has ability to bear individuals as there is no such red colored class names appeared in the inferred class hierarchy. To make this further proved, probe [28] classes are designed and checked with the reasoner as in the Table 1 and results are included according to Figure 4 under result column of the same table.

*Table 1: Probe Classes and Results Obtained for Consistent Checking*

| Probe Class Name | Super class of the Probe Class | | Result |
|---|---|---|---|
| | Super class 1 | Super class 2 | |
| ProbeType1 | Autoimmune | Infectious | Inconsistent |
| ProbeType2 | External | Internal | Inconsistent |
| ProbeType3 | AreaStructure | OrganismStructure | Inconsistent |

ProbeType1 can't exist both under Autoimmune and Infectious as these super-classes are disjoint. ProbeType2 can't exist both under External and Internal as these super-classes are disjoint. ProbeType3 can't exist both under AreaStructure and OrganismStructure as these super-classes are disjoint. Probe classes are removed after consistent checking for disjointed classes has been done. So the proposed Disease ontology passed the consistency checking.

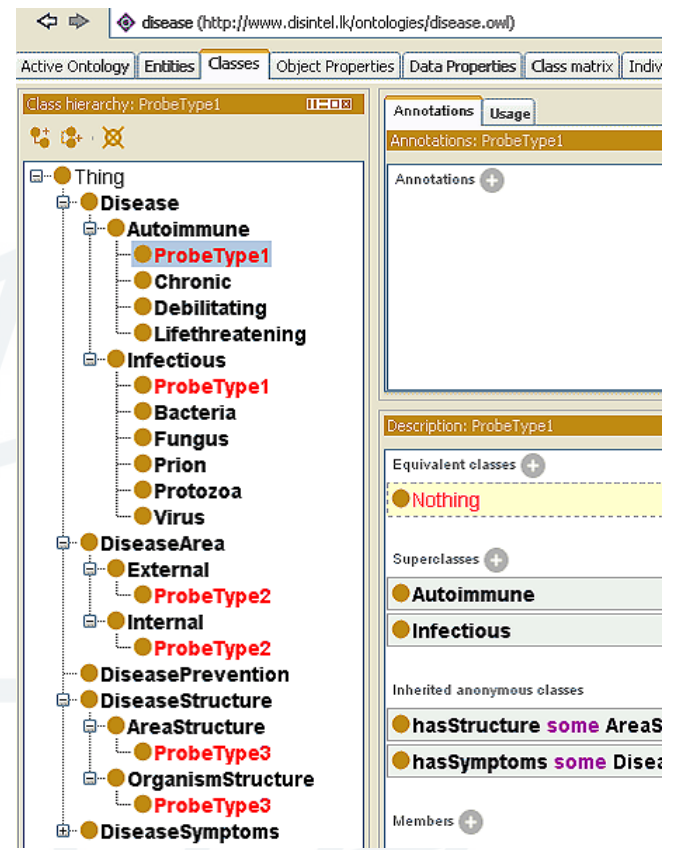

*Figure 4: Probe Classes under Consistent Checking*

The proposed Disease ontology most probably may not be the complete ontology that provides the expected disease intelligence. To make it complete disease ontology, it should be incorporated with millions of data and pieces of information and then test against outputs given by it. This development phase and the testing of it should be validated by the community especially by the domain experts [29]. But in this research, the basic ontology is always checked against the competency questions rather than against requirements specifications because a dynamic requirement specification is expected to be developed with the improvement of the disease ontology with the assistant of the community.

The other way of evaluating this ontology is the use of the web site generated by it. Its basic appearance on the web is depicted in Figure 5. The machine understandable ontology should be able to give correct representation of it as a web through parsers. So the representation should be inspected with a kind of white box testing to evaluate whether the logic behind ontology behave in the correct manner. The white box testing is done by testing the resultant pages appeared when links in the website are clicked and compared with the coding of the ontology at the same time.

The things such as multiple inheritances indirectly associated with the coding related to sub concepts and their way of relation to each other. So the resultant pages are checked with the coding associated with sub concepts and their relations to make correct validation over the ontology. So this white box testing may appear little bit unconventional in this validation of the disease ontology. So each and every linked is checked with resultant pages and coding in the ontology relevant to resultant pages are also checked for consistency of the logic of the disease ontology. Some important results obtained from the white box testing against the proposed Disease web site are shown in Table 2. It shows no inconsistency on the content of the website derived by the proposed Disease ontology.

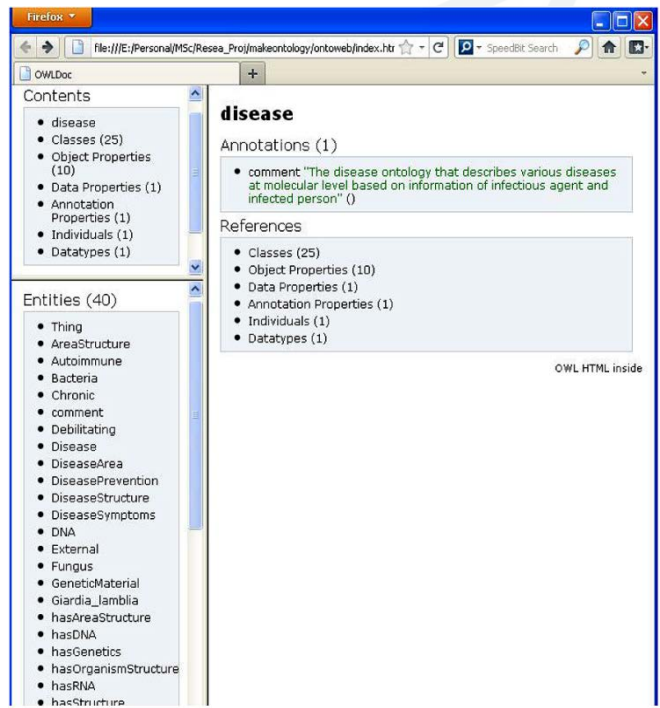

*Figure 5: Basic Structure of the Disease Ontology Website*

*Table 2: Results of White Box Testing Done on the Coding of Proposed Disease Ontology against the Website*

| Content | Result of coding | Result expected from web site |
|---|---|---|
| disease | - | Front page of the web site |
| Classes | 26 classes (including "Thing" class) | 26 classes (including "Thing" class) |
| Object Properties | 10 Object Property | 10 Object Property |
| Data Properties | 1 Data Property | 1 Data Property |
| Annotation Properties | 1 Annotation Property | 1 Annotation Property |
| Individuals | 1 Individual | 1 Individual |
| Data types | 1 Data type | 1 Data type |

Once the objects/individuals have been created, the proposed disease ontology acts as a data repository. As the fed information is stored in logical fashion, machines can interpret them unambiguously and form new relationships which is unforeseen by humans but exist to enlighten the medical field in ways of new drug discoveries and new remedies for curing diseases. For example, when drug details are incorporated into disease it can be expected to have information of drugs relating to disease and the patient. This information reveals how drugs affect the disease and finally to the thousands of patients; giving intelligence for further development of the drugs. Moreover, while the drug business is coming under the purview of Disease Intelligence, it results some kind of business intelligence revolves around it.

The proposed Disease ontology also covers so many areas relating to the disease such as patients' records, clinical trials, micro and macro detail of humans and organisms and so on. The proposed Disease ontology has the advantage of information coming from many areas as well as from many sources in contrast to other ontologies (related to medical field) which have only specified area of consideration. Because of this nature, the disease ontology has advantage of exploring sufficient amount of different links between these entities to make a DIS. Thus, disease intelligence information will be available to researchers, medical practitioners as well as to general public with specificity to their needs through this ontology at the same time.

The proposed Disease ontology has some advantages over well-known ontologies relating to disease. GO which contains three structured controlled ontologies only covers the part of micro level profile of the disease ontology. Also it lacks the clinical aspect of the disease intelligence. Bridging information between diseased and infectious agent is not clearly covered by these ontologies to give clear cut evidence about disease intelligence.

SNOMED CT ontology covers only the clinical aspects of the disease. It does not give clues about special genetic sequence in patients which supports quicker/ slower recovery from cancer. This is because genetic information is not considered in this ontology. Basically it uses patients' clinical records and drugs used for those patients. The intelligence associated with this system basically on how drugs affect the diseased and how patients reactive to some drugs based on different conditions such as sex, age and may be genetics. Micro level analysis of gene in relation to patient and disease is not covered. So this system lacks the following details: the micro and macro level structure of the organisms (if it is an infection) which causes the disease, the DNA / RNA details of the patients etc. So the intelligence regarding to genetic side is not properly covered by this system. The disease intelligence ontology covers all these areas including clinical aspects and so the disease ontology integrates all these aspects. Further the proposed Disease ontology is expected to be able to import SNOMED CT ontology on to the proposed Disease ontology to make the web of data for disease intelligence.

Although the proposed Disease ontology doesn't reach the fine line where ontology diminishes and knowledge base arises, it has formed the basic foundation with core concepts developed with many thoughts that the other developers around the world can easily incorporated into their ontologies and their thoughts to the proposed ontology. As the number of different ontologies which are related to disease ontology are added in exponentially, the task of storing, maintaining and reorganizing them to ensure the successful reuse of ontologies is supposed to be a challenging task.

Other key advantages of the proposed Disease ontology or the cornerstone of the disease intelligence system are use of familiar, local terminology as well as more scientific terminology in combination, support for unanticipated modeling extensions, high degree of automation, high-fidelity integration and mapping with external systems and terminologies and support for accurate answering of expressive queries.

Decade of research and development around Semantic Web technologies still lacks the powerful tools developed for data mining, data management and knowledge discovery from ontologies. User interfaces are still developed with lesser effective and efficient manner, forcing the interface models less attractive for human consumption. So it is necessary to handle the disease ontology within these limitations.

Building an effective Semantic Web for Disease Intelligence would be a long term effort that needs coherent representations along with simple tools to create, publish, query and visualize generic semantic web data.

Another issue which should be discussed in relation to the disease ontology is the wrong data that may give wrong extracted data. The gravity of this issue is mainly based on accuracy of the ontology used in this DIS. Semantics and logical phrases used in this ontology may not cover the wide area of considerations required by such DIS.

## V. CONCLUSIONS

What is expected from the proposed Disease ontology and how it will effectively be evolved to form a kind of intelligence that will pave way to disease intelligence is the main theme of discussion of this research. By building the proposed Disease ontology, it lays foundation for a Disease Intelligence System. It provides best extracted information about rapidly spreading and changing diseases. In addition to that, this information will make information extraction and other natural language processing tools key enablers for the acquisition and use of this semantic information. So it can be used by machines to answer basically the twelve questions regarding human diseases mentioned in the Methodology and Results/ Discussion sections. The proposed Disease ontology should be further developed by the community, once it is available in the web by the means of adding new concepts, refining the existing concepts and adding data/ information to the disease ontology. Until millions of concepts and data are available in the disease ontology, it will not be operated in such a way to give clear and true disease intelligence nature expected by it through the system (web interface) produced by it.